\documentclass[runningheads]{llncs}
\usepackage[T1]{fontenc}
\usepackage{amsmath}
\usepackage{multirow}
\usepackage{hyperref}
\usepackage{amssymb}

\usepackage{graphicx}
\usepackage{booktabs} 
\usepackage{pifont}
\begin{document}
\title{Supervised Contrastive Machine Unlearning of Background Bias in Sonar Image classification with Fine-Grained Explainable AI}
%
%
\author{Kamal Basha~S~$^{1}$ and Athira~Nambiar$^{1}$}%
\authorrunning{F. Author et al.}
%
\institute{Department of Computational Intelligence,\\
Faculty of Engineering and Technology,\\
SRM Institute of Science and Technology\\
Kattankulathur, Tamil Nadu, 603203, India\\ 
\email{c58527@srmist.edu.in, athiram@srmist.edu.in}}
\maketitle              
\begin{abstract}
Acoustic sonar image analysis plays a critical role in object detection and classification, with applications in both civilian and defense domains. Despite the availability of real and synthetic datasets, existing AI models that achieve high accuracy often over-rely on seafloor features, leading to poor generalization. To mitigate this issue, we propose a novel framework that integrates two key modules: (i) a Targeted Contrastive Unlearning (TCU) module, which extends the traditional triplet loss to reduce seafloor-induced background bias and improve generalization, and (ii) the Unlearn to Explain Sonar Framework (UESF), which provides visual insights into what the model has deliberately forgotten while adapting the LIME explainer to generate more faithful and localized attributions for unlearning evaluation. Extensive experiments across both real and synthetic sonar datasets validate our approach, demonstrating significant improvements in unlearning effectiveness, model robustness, and interpretability.

\keywords{Sonar image analysis \and Contrastive unlearning \and classification \and Seafloor bias \and Explainable AI}

\end{abstract}
%
%
%

\section{Introduction}

Sonar image classification has emerged as a critical task for underwater surveillance and  maritime security~\cite{chai2023deep}. With the advancement of machine learning and deep learning techniques, several frameworks have been proposed to automatically identify underwater objects such as mines, wrecks, and ships~\cite{chungath2023transfer}. However, despite high accuracy on benchmark datasets, these models often exhibit a critical weakness: \textbf{over-reliance on the seafloor}. As a recurring, low-variance background, the seafloor dominates most sonar images, causing models to latch onto background cues rather than object-specific features. This bias leads to poor generalization and provides false positive, especially in real-world deployment scenarios.

The field of \textbf{machine unlearning}~\cite{nguyen2022survey} has recently gained traction, primarily within the context of privacy-preserving systems where specific data samples are intentionally forgotten from trained models~\cite{li2025machine,chen2025survey}. These techniques have largely focused on unlearning sensitive user data or removing information to meet legal constraints such as General Data Protection Regulation (GDPR). However, the application of machine unlearning to address semantic biases such as background dependence in sonar images remains unexplored. In parallel \textbf{Contrastive learning}~\cite{hu2024comprehensive} has emerged as a powerful tool for representation learning capabilities across vision tasks, especially in self-supervised settings. While it has been employed for sonar feature discrimination, its adaptation for targeted unlearning in supervised settings has not been studied. Such an approach could explicitly suppress background correlations while preserving object-level semantics.

\textbf{Explainable Artificial Intelligence (XAI)}~\cite{tjoa2020survey} aims to transition AI systems from black-box to white-box models by providing transparent and human-interpretable justifications for their predictions. However, model-agnostic techniques~\cite{jiarpakdee2020empirical} such as LIME often highlight irrelevant regions (e.g., the seafloor in sonar imagery) as dominant contributors, leading to misleading explanations and revealing hidden biases in the model.

In order to address the challenges of seafloor-induced background bias and the lack of interpretable evidence for what has truly been forgotten by the model, a unified framework is introduced that combines supervised contrastive unlearning with fine-grained explainability. At the core of this framework lies the \textbf{Targeted Contrastive Unlearning (TCU)} module, in which a modified triplet loss formulation is employed. Anchor and positive samples are drawn from the same object class, while the negative sample is consistently selected from the \textit{seafloor} class. This targeted sampling explicitly directs the model to suppress background-related representations and focus on learning semantic features of foreground objects. The final objective integrates this triplet loss with a contrastive entropy term, both optimized over the feature embeddings produced by a shared backbone. To quantify and visualize the unlearning effect, a \textbf{Unlearn to Explain Sonar Framework (UESF)} is introduced. By applying LIME to both the baseline and unlearned models, the explainer captures differential attribution maps, enabling a direct comparison of attention regions before and after unlearning. This facilitates a measurable and interpretable assessment of whether the background cues have been effectively removed, and whether the explainer itself has shifted focus toward meaningful object regions.

This work marks the first integration of supervised contrastive unlearning and explainer fine-tuning in sonar image analysis. By simultaneously mitigating background bias and enhancing the trustworthiness of model explanations, the proposed framework establishes a new direction toward transparent, deployable, and bias-resilient sonar AI systems.

The key contributions of this work are summarized as follows:
\begin{itemize}
    \item Proposal of a supervised contrastive unlearning strategy through the \textbf{Targeted Contrastive Unlearning (TCU)} module, explicitly designed to mitigate seafloor bias in sonar imagery.
    
    \item Incorporation of a novel \textbf{modified triplet-based supervised learning} formulation to enhance discriminative feature learning while promoting selective forgetting.
    
    \item Introduction of an Unlearn to \textbf{Explain Sonar Framework (UESF) module}, enabling both visualization and quantification of the forgetting process by tracking reduced background relevance.
    
    \item \textbf{Fine-tuning LIME explainer} into the unlearning framework to provide more precise, object-centric, and semantically relevant explanations.
    
    \item Extensive validation on real and synthetic sonar datasets, demonstrating effective mitigation of seafloor dependence and improved target-centric detection performance.
\end{itemize}

\section{Related Works}
\subsection{Sonar Image Classification}
Deep learning in sonar imagery has advanced through transfer learning and few-shot learning to address limited and noisy data \cite{chungath2023transfer}. Synthetic datasets such as S3Simulator and its enhanced variant S3Simulator+ provide realistic side-scan sonar (SSS) data with physics-informed noise for robust model development \cite{kamal2024s3simulator,Basha2025ANC}. Domain generalization techniques, such as Syn2Real using diffusion models, improve mine-like object detection by transferring from synthetic to real data, yielding substantial precision gains \cite{10919029}. Architectural strategies like AW-CNN enhance feature extraction \cite{wang2019underwater}, while Feature Contrastive Transfer Learning (FCTL) combines transfer and contrastive learning to address few-shot, long-tailed classification challenges \cite{bai2025feature}. Additional approaches leverage context-adaptive fusion of shadow and highlight features for improved robustness against noise and occlusion in sonar classification \cite{Basha2025ANC}. Contrastive domain adaptation further improves generalized zero-shot classification in unseen sonar categories \cite{jia2025contrastive}. Machine unlearning methods remove specific training data to meet privacy requirements such as GDPR \cite{zhang2023review}. Contrastive unlearning refines this by separating unlearning samples in representation space while retaining performance on remaining data \cite{kyu2024contrastive}.

\subsection{Explainable AI for Fine-Grained Analysis}
Explainability in fine-grained vision models has advanced through prompt-based attention mechanisms such as Prompt-CAM, which enables localized feature attribution in vision transformers \cite{chowdhury2025prompt}. In sonar imaging, interpretability has been improved by semi-supervised salient object detection that combines uncertainty estimation with contrastive learning to better highlight target regions \cite{yuan2025semi}. Feature attribution for underwater mine warfare has been explored through perturbation-based XAI methods, which outperform backpropagation approaches for grayscale sonar data and yield feature maps consistent with human operator reasoning \cite{richard2024explainable}. Context-adaptive fusion of shadow and highlight regions, coupled with a region-aware denoising model guided by explainability, has further enhanced classification robustness and interpretability in noisy underwater environments \cite{Basha2025ANC}.

Despite advances in classification, unlearning, and explainability, background bias in sonar imagery remains under-addressed, and bias-focused interpretability after unlearning is rarely explored. To the best of our knowledge, this work marks the first attempt to combine targeted contrastive unlearning with explainability to simultaneously mitigate background bias and provide bias-aware interpretations.

\section{Methodology}

\begin{figure}[ht]
    \centering
    \includegraphics[width=\textwidth]{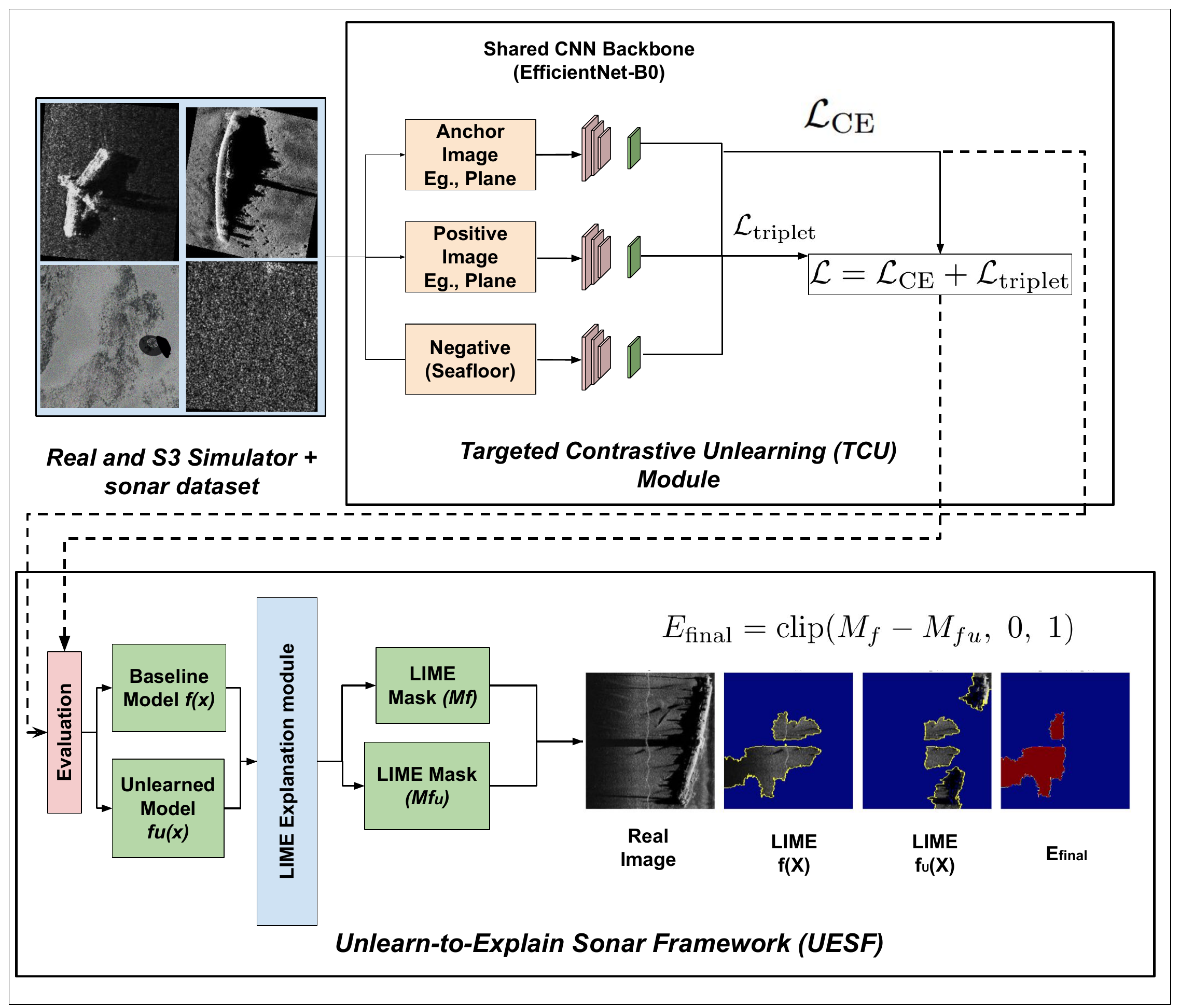} 
   \caption{Proposed architecture integrating contrastive machine unlearning with explainability. The pipeline includes a baseline classifier $f(x)$, contrastive unlearning model $f_u(x)$, and the machine explainer module to visualize and compare saliency maps, ensuring the model effectively forgets background (seafloor) cues while retaining object-specific features.
}
    \label{fig:archi}
\end{figure}

This section outlines the proposed framework, illustrated in Fig.~\ref{fig:archi}, and its key components for robust sonar image classification and interpretability. Section~\ref{dataset} introduces the dataset and preprocessing strategies. Section~\ref{TCU} presents the \textit{Targeted Contrastive Unlearning (TCU)} module, designed to selectively forget seafloor-related features and enhance generalization. Section~\ref{loss} details the associated loss functions and optimization strategies. Finally, Section~\ref{UESF} describes the \textit{Unlearn-to-Explain Sonar Framework (UESF)}, which integrates unlearning with explainability to deliver interpretable insights into sonar-based decision-making.

\subsection{Dataset and Preprocessing}
\label{dataset}
\begin{figure}[ht]
    \centering
    \includegraphics[width=0.8\linewidth]{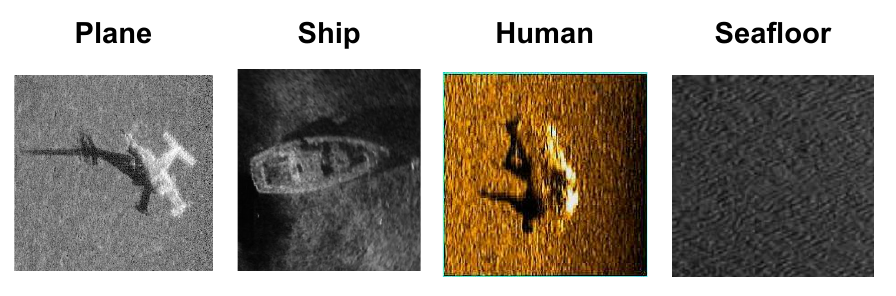}
    \caption{Sample of real sonar images used in this study.}
    \label{fig:real_images}
\end{figure}

\begin{figure}[ht]
    \centering
    \includegraphics[width=0.6\linewidth]{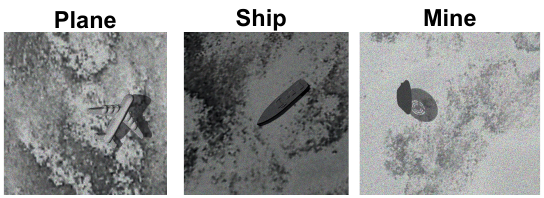}
    \caption{Sample of synthetic sonar images from the S3Simulator+ dataset.}
    \label{fig:s3sim_images}
\end{figure}

The dataset used in this study consists of both real sonar images as shown in Fig.\ref{fig:real_images} and simulated sonar images i.e., S3Simulator+ as show in Fig.~\ref{fig:s3sim_images}. Specifically, the SeabedObjects-KLSG~\cite{huo2020underwater}, Sonar Common Target Detection (SCTD) dataset~\cite{zhang2021self}, and S3Simulator+~\cite{kamal2024s3simulator,basha2025novel} datasets were employed. In total, approximately 2,500 images were collected, covering five categories: ship, plane, mine, human, and seafloor. The dataset was partitioned into 70\% for training, 15\% for validation, and 15\% for testing to ensure a balanced evaluation.

Each image was preprocessed by resizing to a fixed resolution of 224$\times$224 pixels, followed by normalization to standardize pixel intensity values. To enhance the robustness and generalization capability of the models, various data augmentation techniques~\cite{shorten2019survey} such as rotation, flipping, and scaling were applied during training.


\subsection{Targeted Contrastive Unlearning (TCU) Module}
\label{TCU}

To mitigate background bias in sonar image classification, a supervised contrastive unlearning module is incorporated. The classification task is reformulated as a triplet-based~\cite{schroff2015facenet} representation learning problem. For each anchor sample $x_i^a$, a positive sample $x_i^p$ is drawn from the same semantic class, while a negative sample $x_i^n$ is consistently chosen from the background class (\textit{seafloor}). This construction deviates from conventional triplet learning, where negatives are selected from arbitrary classes other than the anchor's. By fixing the background as the negative reference, the encoder is explicitly guided to suppress background-specific features and focus on discriminative object semantics.

The triplet unlearning loss is formulated as:
\begin{equation}
\mathcal{L}_{\text{triplet}} = \sum_{i=1}^{N} \max \Big( 0, \ \| f(x_i^a) - f(x_i^p) \|_2^2 - \| f(x_i^a) - f(x_i^n) \|_2^2 + \alpha \Big),
\end{equation}

where $f(\cdot)$ denotes the encoder network, $\alpha$ is the margin hyperparameter, and $[\cdot]_+$ is implemented through the hinge function. This objective enforces compactness between anchor–positive embeddings while maximizing separation from background embeddings, thereby attenuating bias.

\subsection{Loss Function and Optimization}
\label{loss}

The overall training objective integrates supervised classification with targeted contrastive unlearning. Formally, the total loss is given by
\begin{equation}
\mathcal{L}_{\text{total}} 
= \mathcal{L}_{\text{CE}} 
+ \lambda \, \mathcal{L}_{\text{triplet}},
\end{equation}
where $\mathcal{L}_{\text{CE}}$ denotes the standard cross-entropy loss for class prediction, and $\mathcal{L}_{\text{triplet}}$ represents the triplet loss that encourages discriminative and background-invariant representations. The trade-off parameter $\lambda > 0$ is tuned empirically via grid search.


\subsection{Unlearn-to-Explain Sonar Framework (UESF)}
\label{UESF}

To interpret and verify the effect of unlearning, the Unlearn-to-Explain Sonar Framework (UESF) is employed. This module evaluates whether background cues are suppressed during the decision-making process. Local Interpretable Model-Agnostic Explanations (LIME)~\cite{ribeiro2016should} is applied to generate saliency masks for the baseline model $f(x)$ and the unlearned model $f_u(x)$.

The explanation difference score is computed as:
\begin{equation}
E_{\text{Final}} = \mathrm{clip}\!\left(M_f - M_{fu},\,0,\,1\right),
\end{equation}
where $M_f$ and $M_{fu}$ denote the binary masks highlighting background-focused regions for the baseline and unlearned models, respectively. A higher $E_{\text{Final}}$ corresponds to greater divergence between the models’ reliance on background cues. A consistent reduction of background attribution in $f_u(x)$ indicates effective unlearning, while simultaneously improving interpretability by revealing which features were suppressed.

\section{Experimental Setup}
\subsection{Evaluation metrics}

In the evaluation of sonar image classification, the classification report~\cite{powers2020evaluation} consisting of accuracy, recall, precision, and F1-score was employed to assess quantitative performance. Confusion Matrix were generated to illustrate class-wise prediction strengths and misclassifications. To visualize feature separability, t-SNE plots~\cite{maaten2008visualizing} were utilized for dimensionality reduction and representation of learned embeddings. Furthermore, interpretability was addressed through LIME-based visual explanations, enabling the comparison of baseline and unlearned models. The differences highlighted by the Unlearn to Explain Sonar Framework (UESF) were emphasized to demonstrate the extent of knowledge that was intentionally forgotten.

\subsection{Implementation Details}

For the baseline model, EfficientNet-B0 pre-trained on ImageNet was employed as the backbone to perform sonar image classification, given its proven effectiveness in underwater object recognition~\cite{arjun2024unveiling}. The dataset, comprising both real and synthetic sonar images, was split into training, validation, and testing sets in a 70/15/15 ratio. A batch size of 16 was adopted using PyTorch’s DataLoader, with shuffling enabled for training while keeping validation and test sets sequential. The model was trained for 20 epochs with the Adam optimizer (learning rate = 0.0001) and a cross-entropy loss function. Model checkpointing was utilized to preserve the best-performing model based on validation accuracy.

For the unlearning framework, the proposed \textit{Targeted Contrastive Unlearning (TCU)} module was implemented through a triplet-based training pipeline. A custom designed to generate anchor, positive, and negative samples, where the seafloor class served as the primary negative reference to mitigate background bias. The training objective combined cross-entropy loss with a triplet margin loss (margin = 1.0, $p=2$). Optimization was performed using Adam (learning rate = 0.0001) for 20 epochs. To ensure stability, debugging checks such as sample inspection, dataloader verification, and forward-pass validation were integrated during training.

To assess interpretability, the \textit{Unlearn to Explain Sonar Framework (UESF)} was employed. Specifically, the LIME image explainer was fine-tuned to highlight discriminative cues for both the baseline model $f(x)$ and the unlearned model $f_u(x)$. Each sonar input image was resized to $224 \times 224$ pixels and normalized prior to inference. An explanation-difference heatmap $E_{\text{final}}$ was computed, quantifying the reduction of background-focused regions and revealing what knowledge was effectively forgotten through unlearning.

The entire implementation was carried out in PyTorch with Torchvision utilities for model adaptation, and LIME with scikit-image for explainability. Experiments were executed in a Python 3.10 environment on CUDA-enabled GPUs, ensuring efficient training and evaluation across sonar imaging scenarios. Training on an NVIDIA RTX 3060 GPU required approximately 22 minutes for 20 epochs, confirming the framework’s efficiency.

\section{Experimental Results}
In Section \ref{sec:qanti}, quantitative results are presented through the classification report, confusion matrix, and t-SNE plots. Section \ref{sec:quali} provides qualitative analysis, where LIME explanations are compared across the baseline and unlearned models.
\subsection{Quantitative Results}
\label{sec:qanti}
From Table~\ref{tab:classification_report}, it can be seen that although both the baseline and unlearned EfficientNet-B0 achieve comparable overall accuracy ($0.99$), subtle yet meaningful differences emerge. The baseline model shows slightly lower recall for the ``Plane'' class ($0.96$), suggesting occasional confusion with seafloor textures. In contrast, the unlearned model improves recall for ``Plane'' ($0.99$) and provides more balanced precision–recall tradeoffs for ``Ship,'' indicating enhanced robustness against background bias. These results confirm that Targeted Contrastive Unlearning (TCU) preserves classification performance while mitigating over-reliance on seafloor features, thereby improving generalization without sacrificing accuracy.

\begin{table}[htbp]
\centering
\caption{Classification performance comparison of baseline EfficientNet-B0 and Unlearned EfficientNet-B0 models.}
\label{tab:classification_report}
\begin{tabular}{lccccc}
\hline
\textbf{Class} & \textbf{Accuracy} & \textbf{Precision} & \textbf{Recall} & \textbf{F1-Score} & \textbf{Support} \\
\hline
\multicolumn{6}{l}{\textit{Baseline (EfficientNet-B0)}} \\
Human     & 0.99 & 1.00 & 0.99 & 0.99 & 97 \\
Mine      & 1.00 & 1.00 & 1.00 & 1.00 & 80 \\
Plane     & 0.97 & 0.99 & 0.96 & 0.98 & 84 \\
Seafloor  & 1.00 & 1.00 & 1.00 & 1.00 & 101 \\
Ship      & 0.98 & 0.96 & 0.99 & 0.97 & 89 \\
\textbf{Overall / Avg. Accuracy} & \textbf{0.99} & \multicolumn{4}{c}{} \\
\hline
\multicolumn{6}{l}{\textit{Unlearned EfficientNet-B0}} \\
Human     & 1.00 & 1.00 & 1.00 & 1.00 & 97 \\
Mine      & 1.00 & 1.00 & 1.00 & 1.00 & 80 \\
Plane     & 0.98 & 0.98 & 0.99 & 0.98 & 84 \\
Seafloor  & 1.00 & 1.00 & 1.00 & 1.00 & 101 \\
Ship      & 0.99 & 0.99 & 0.98 & 0.98 & 89 \\
\textbf{Overall / Avg. Accuracy} & \textbf{0.99} & \multicolumn{4}{c}{} \\
\hline
\end{tabular}
\end{table}

\begin{figure}[ht]
    \centering
    \includegraphics[width=\textwidth]{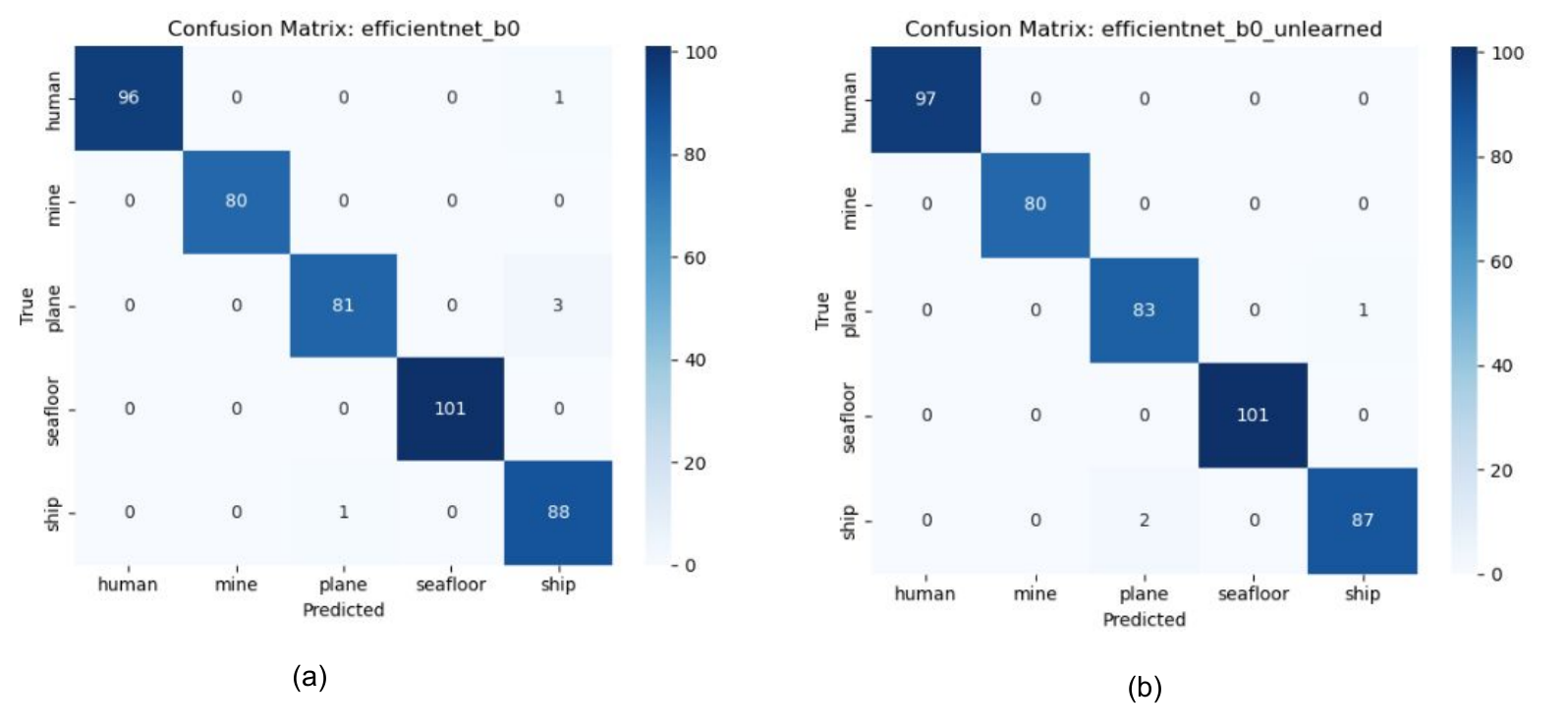} 
   \caption{Confusion matrix showing class-wise prediction accuracy before and after applying the unlearning framework.
}
    \label{fig:confusion_matrix}
\end{figure}

From Fig.~\ref{fig:confusion_matrix}, both the baseline and unlearned EfficientNet-B0 models achieve consistently high classification accuracy across all classes. However, the baseline model exhibits occasional misclassifications of ``Plane'' as ``Ship'' and vice versa, reflecting sensitivity to background textures. In contrast, the unlearned model reduces such errors, particularly improving recognition of the ``Plane'' class (83 correct vs. 81 in baseline) while maintaining perfect classification of ``Seafloor'' and ``Mine.'' These results further support that Targeted Contrastive Unlearning (TCU) enhances robustness to seabed bias without degrading overall accuracy.

\begin{figure}[htbp]
    \centering
    \includegraphics[width=0.95\linewidth]{images/t-SNE.pdf}
    \caption{t-SNE visualizations of feature embeddings after PCA$\rightarrow$t-SNE. 
    (\textbf{a}) Baseline model: seabed samples (purple) are entangled with ship (red), mine (orange), and plane (green), indicating strong seabed bias and reduced class separability. 
    (\textbf{b}) Unlearned model: seabed forms a distinct and isolated cluster, clearly separated from human (blue), mine (orange), ship (red), and plane (green), demonstrating the effectiveness of the unlearning process in mitigating seabed bias and improving class-specific representations (better viewed in colour).}
    \label{fig:tsne}
\end{figure}

Fig.~\ref{fig:tsne} compares the t-SNE embeddings of the baseline and unlearned models. 
In the baseline case, seabed features dominate the representation space, causing overlap with ship, mine, and plane classes, which highlights the presence of seabed bias. 
In contrast, the unlearned model produces a distinct separation between seabed and object classes, confirming that the unlearning process effectively mitigates seabed bias and enhances class separability.

\subsection*{Comparative Analysis with Related Works}

In contrast, the proposed \textbf{Targeted Contrastive Unlearning (TCU)} module and \textbf{Unlearn-to-Explain Sonar Framework (UESF)} constitute the first unified attempt to jointly perform bias mitigation and explainability in sonar imagery. By integrating supervised contrastive unlearning with LIME-based differential visualization, our framework maintains competitive accuracy (0.99) while deliberately reducing background dependence and enhancing interpretability. Table~\ref{tab:comparison_related_work_full} summarizes the performance of representative sonar image classification frameworks alongside our proposed approach. This highlights the quantitative gains and the novel contributions of UESF in terms of explainability and bias mitigation.

\begin{table}[h!]
\centering
\caption{Comparison of sonar image classification, explainability, and bias mitigation frameworks with the proposed TCU + UESF approach.}
\label{tab:comparison_related_work_full}
\begin{tabular}{|p{2cm}|p{3cm}|p{3cm}|p{3cm}|}
\hline
\textbf{Method} & \textbf{Focus} & \textbf{Explainability Support} & \textbf{Bias Mitigation} \\ \hline
AW-CNN~\cite{wang2019underwater} & CNN-based sonar classification & \texttimes & \texttimes \\ \hline
FCTL~\cite{bai2025feature} & Contrastive transfer learning & \texttimes & \texttimes \\ \hline
Syn2Real~\cite{10919029} & Domain generalization via diffusion & \texttimes & \texttimes \\ \hline
Context-Adaptive Fusion~\cite{Basha2025ANC} & Shadow–highlight feature fusion & \checkmark & \texttimes \\ \hline
Contrastive Domain Adaptation~\cite{jia2025contrastive} & Zero-shot / generalized classification & \texttimes & \texttimes \\ \hline
Machine Unlearning~\cite{zhang2023review} & Privacy-preserving removal of data & \texttimes & \checkmark \\ \hline
Contrastive Unlearning~\cite{kyu2024contrastive} & Contrastive representation unlearning & \texttimes & \checkmark \\ \hline
Prompt-CAM~\cite{chowdhury2025prompt} & Vision-transformer explainability & \checkmark & \texttimes \\ \hline
Semi-supervised Saliency~\cite{yuan2025semi} & Highlighting target regions in sonar & \texttimes & \texttimes \\ \hline
Perturbation-based XAI~\cite{richard2024explainable} & Feature attribution for sonar & \checkmark & \texttimes \\ \hline
\textbf{Proposed TCU + UESF} & Contrastive unlearning + explainability & \checkmark~(LIME-based differential visualization) & \checkmark~(Seafloor bias unlearning) \\ \hline
\end{tabular}
\end{table}

\subsection{Qualitative Results}
\label{sec:quali}

\begin{figure}[ht]
    \centering
    \includegraphics[width=\textwidth]{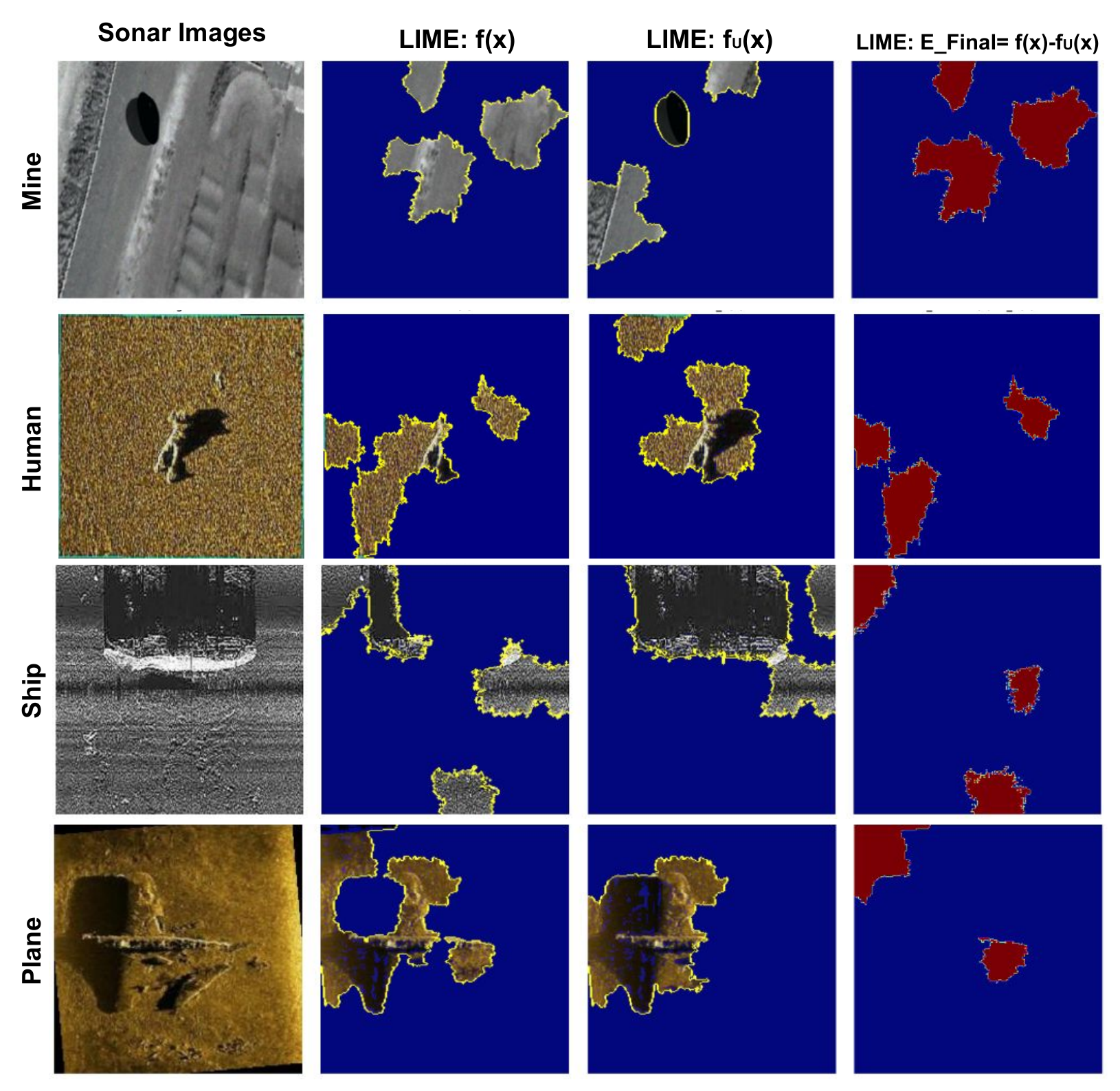} 
   \caption{
Qualitative comparison of LIME explanations before and after unlearning. 
$E_{\text{final}} = f(x) - f_u(x)$ highlights regions forgotten by the model, revealing reduced background bias.
}
    \label{fig:lime_qualitative}
\end{figure}

Fig.~\ref{fig:lime_qualitative} illustrates the impact of the proposed framework on LIME explanations for sonar images. 
In the baseline model ($f(x)$), the highlighted regions show strong dependence on seabed textures, confirming the presence of seafloor bias that limits generalizability. 
After applying Targeted Contrastive Unlearning (TCU), the unlearned model ($f_u(x)$) attends more precisely to object regions while discarding irrelevant background patterns. 
The difference map $E_{\text{final}} = f(x) - f_u(x)$, enabled by the Unlearn to Explain Sonar Framework (UESF), clearly isolates the forgotten seabed features, validating both the effectiveness of unlearning and the improved faithfulness of the explanations. 
These results demonstrate that our framework successfully mitigates seafloor bias and enhances interpretability, ultimately supporting more reliable sonar-based object detection.

\section{Conclusion and Future works}
In this work, we introduced a novel Supervised Contrastive Machine Unlearning framework for underwater sonar image analysis, addressing the critical challenge of background bias. By systematically combining real and simulated data, we proposed a Targeted Contrastive Unlearning (TCU) module, where contrastive unlearning was applied within the classification pipeline to mitigate bias and enhance robustness. Furthermore, the Unlearn-to-Explain Sonar Framework (UESF) was developed to provide interpretability, highlighting what the model has intentionally forgotten through visual explanations. This dual perspective of performance improvement and interpretability assurance establishes a strong foundation for deploying trustworthy sonar-based decision systems. Future work will focus on extending the framework’s flexibility by incorporating advanced explainability techniques (e.g., SHAP, Integrated Gradients, and counterfactual explanations) to capture more nuanced aspects of model reasoning and XAI quantitative metrics. Additionally, we plan to explore Vision Transformers and hybrid attention mechanisms to model deeper representational structures in sonar imagery, enabling more robust unlearning and improved generalization across diverse underwater environments.

\bibliographystyle{splncs04}
\bibliography{mybibliography}

\end{document}